\pdfoutput=1

\documentclass[11pt]{article}

\usepackage{acl}
\usepackage{times}
\usepackage{graphicx}
\usepackage{latexsym}
\usepackage{amsmath}
\usepackage{amssymb} 
\usepackage{algorithm}
\usepackage{algpseudocode} 
\usepackage{xspace}
\usepackage{multirow}
\usepackage{booktabs}
\usepackage[T1]{fontenc}

\usepackage[utf8]{inputenc}

\usepackage{microtype}

\usepackage{inconsolata}
\usepackage{color}

\newcommand{\proposed}{RoleFact\xspace}
\newcommand{\bluebold}[1]{\textbf{\textcolor{blue}{#1}}}
\newcommand{\redul}[1]{\textcolor{red}{\underline{#1}}}

\title{Mitigating Hallucination in Fictional Character Role-Play}

\author{Nafis Sadeq$^{ 1}$, Zhouhang Xie$^{ 1}$, Byungkyu Kang$^{ 2}$ \\
{\bf Prarit Lamba $^{ 2}$, Xiang Gao$^{ 2}$, Julian McAuley$^{ 1}$} \\
   UC San Diego$^{1}$ \and Intuit$^{2}$ \\
    \texttt{\{nsadeq,zhx022,jmcauley\}@ucsd.edu} \\
    \texttt{\{Jay\_Kang,Prarit\_Lamba,xiang\_gao\}@intuit.com}}

\begin{document}
\maketitle
\begin{abstract}
Role-playing has wide-ranging applications in customer support, embodied agents, and computational social science. The influence of parametric world knowledge of large language models (LLMs) often causes role-playing characters to act out of character and to hallucinate about things outside the scope of their knowledge. In this work, we focus on the evaluation and mitigation of hallucination in fictional character role-play. We introduce a dataset with over 2,000 characters and 72,000 interviews, including 18,000 adversarial questions. We propose \proposed, a role-playing method that mitigates hallucination by modulating the influence of parametric knowledge using a pre-calibrated confidence threshold. Experiments show that the proposed method improves the factual precision of generated responses by 18\% for adversarial questions with a 44\% reduction in temporal hallucination for time-sensitive interviews. The code and the dataset are available at \url{https://github.com/NafisSadeq/rolefact.git}.
\end{abstract}
\section{Introduction}
The performance of LLMs on simple downstream tasks such as question answering~\cite{rajpurkar-etal-2018-know}, text classification~\cite{wang-etal-2018-glue}, natural language understanding~\cite{Hendrycks2020mmlu}, etc. has improved significantly~\cite{brown2020gpt3,bubeck2023sparks}, prompting the question whether they can perform more complex agent level tasks such human simulacra~\cite{park2023generative}, tool use~\cite{schick2023toolformer}, and role-playing~\cite{shao2023characterllm}. Role-playing aims to enable an AI system to impersonate a domain expert~\cite{xu2023expertprompting}, companion for psychological support~\cite{tu2023characterchat}, or a fictional character~\cite{wang2024rolellm} by mimicking their knowledge, personality traits, and speaker style. Applications of role-playing include customer support~\cite{Soni_2023}, embodied agents~\cite{huang2022embodied}, and computational social science~\cite{zhou2024sotopia}. While LLMs like GPT-4~\cite{openai2024gpt4}, Gemini~\cite{geminiteam2024gemini}, Llama-3~\cite{meta-llama3} demonstrate some role-playing capabilities, the role-playing agent often hallucinates when asked about things that should be outside of their knowledge base~\cite{shao2023characterllm}.

Hallucination remains a challenge for most LLM-based systems~\cite{Guan_Liu_AAAI_2024}. While retrieval augmented generation (RAG)~\cite{karpukhin2020dense,lewis2020retrieval} may mitigate some hallucination induced by outdated knowledge in language models~\cite{shuster2021retrieval}, it cannot guarantee that most claims made by the system are supported by a non-parametric knowledge base~\cite{asai2023selfrag}. Learning to improve character role-playing methods so that the responses are verifiable by a story-specific script serves as a case study that can help mitigate hallucination for the RAG paradigm in general. 

\begin{figure}
  \centering
  \includegraphics[width=\columnwidth]{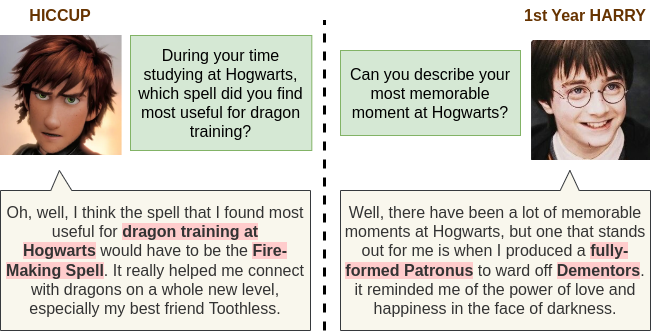}
 \caption{Example of cross-universe hallucination (Hiccup should not answer questions about Hogwarts) and temporal hallucination (Harry should not talk about the Patronus charm in his first year) in character role-play.}
 \vspace{-1em}
  \label{fig:intro_rolefact}
\end{figure}

\begin{figure}
\centering
\includegraphics[width=0.95\columnwidth]{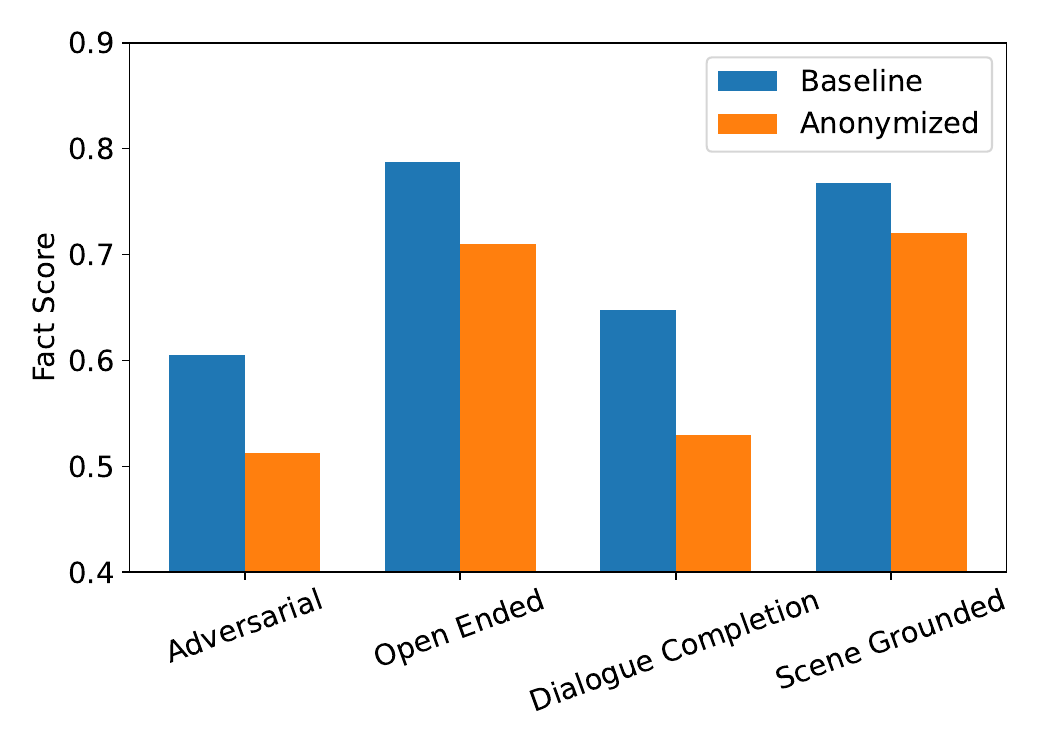}
\caption{Factual precision degrades when we minimize parametric knowledge by anonymizing the prompts.}
\label{fig:anonymous}
\vspace{-1em}
\end{figure}

LLMs are pretrained on a large corpus of internet data and their vast parametric knowledge helps them perform a diverse range of tasks~\cite{radford2019language,brown2020gpt3}. However, the parametric knowledge of LLMs makes the role-playing character vulnerable to out-of-character behaviors and hallucinations~\cite{shao2023characterllm}. A role-playing character may hallucinate about things that are outside of their story-line (e.g. Hiccup from `How to Train Your Dragon' talking about learning a fire-making spell at Hogwarts) or demonstrate knowledge of events ahead of time (Harry Potter as a first-year Hogwarts student talks about producing the Patronus charm). A role-playing character should be able to demonstrate a lack of knowledge about things that are not supported by a story-specific script or knowledge outside of a relevant temporal window in case of time-sensitive role-play. One approach to achieve this is to strictly remove parametric knowledge from the response and solely rely on non-parametric retrieved knowledge. However, this approach degrades informativeness and it can even hurt factual precision. This is shown in Figure~\ref{fig:anonymous} where we anonymized the prompts to minimize the role of parametric knowledge. We argue that an ideal role-playing approach should modulate the influence of parametric knowledge to find the balance between factuality and informativeness. 

Prior methods of role-play mostly rely on role-profile prompting~\cite{wang2024rolellm} and knowledge retrieval in a time-agnostic fashion~\cite{li2023chatharuhi}. These approaches heavily depend on parametric knowledge of LLMs which leads to hallucinations and poor performance for less popular characters (shown in Figure~\ref{fig:long_tail_factual_precision}). Currently, a major challenge for studying hallucination in character role-play is the lack of a suitable dataset. The lack of a dataset forces researchers to rely on a rating-based evaluation of hallucination~\cite{shao2023characterllm,lu-etal-2024-large} which is highly subjective. Prior datasets in the role-play domain are insufficient for studying time-sensitive character role-play or hallucination for less popular characters. Our contributions are as follows:

\begin{figure}
\centering
\includegraphics[width=0.95\columnwidth]{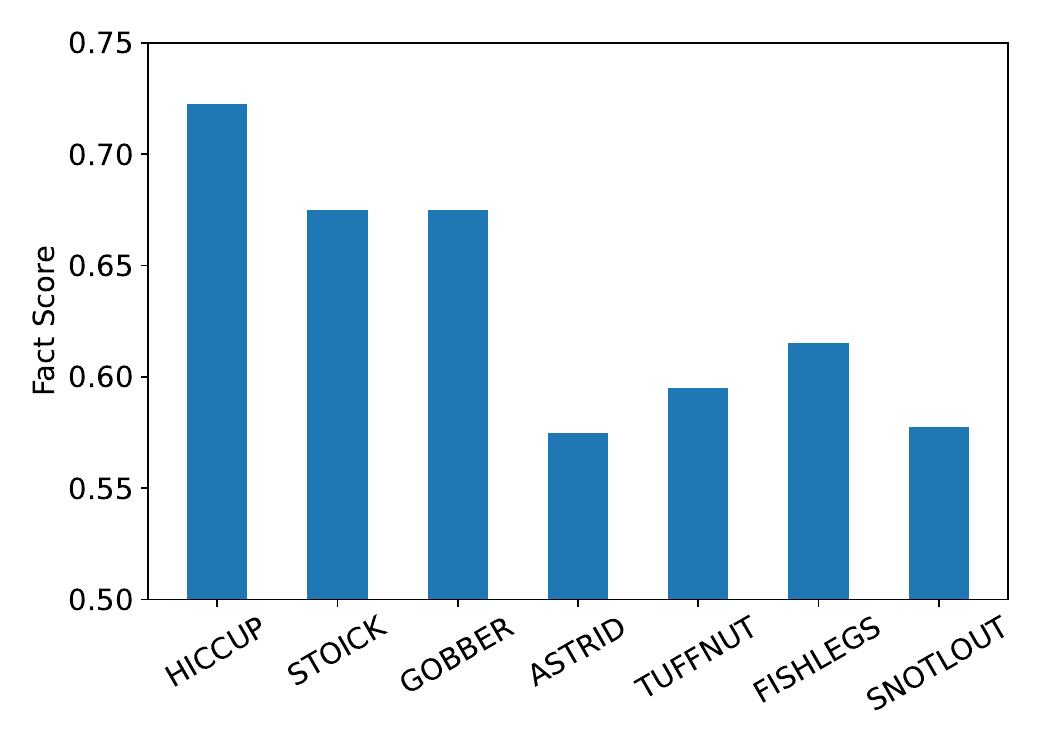}
\caption{Factual precision degrades with decreasing character popularity (left to right), shown for characters associated with the `How to Train Your Dragon' series.}
\label{fig:long_tail_factual_precision}
\vspace{-1em}
\end{figure}

\begin{itemize}
    \item We release a dataset for Script Grounded Character Role-play (SGR) that includes more than 2,000 characters, and 72,000 interviews. With script-specific knowledge and 4 diverse tasks, the dataset aims to enable a systematic study of character hallucinations. To the best of our knowledge, this is the largest and the most diverse dataset that facilitates the study of temporal hallucination and hallucination for less popular characters. Script-specific knowledge aligned with the given tasks also facilitates a more nuanced evaluation of hallucination based on atomic fact decomposition~\cite{min2023factscore}. 
    \item  We propose \proposed, a role-playing method that modulates the influence of parametric knowledge of LLMs on character responses. \proposed improves factual precision by 18\% for adversarial interviews, reduces temporal hallucination by 44\% for time-sensitive interviews, and improves factual precision by 23\% for less popular characters.
\end{itemize}
\section{Methodology}
 \begin{figure*}
    \centering
	\includegraphics[width=0.95\linewidth]{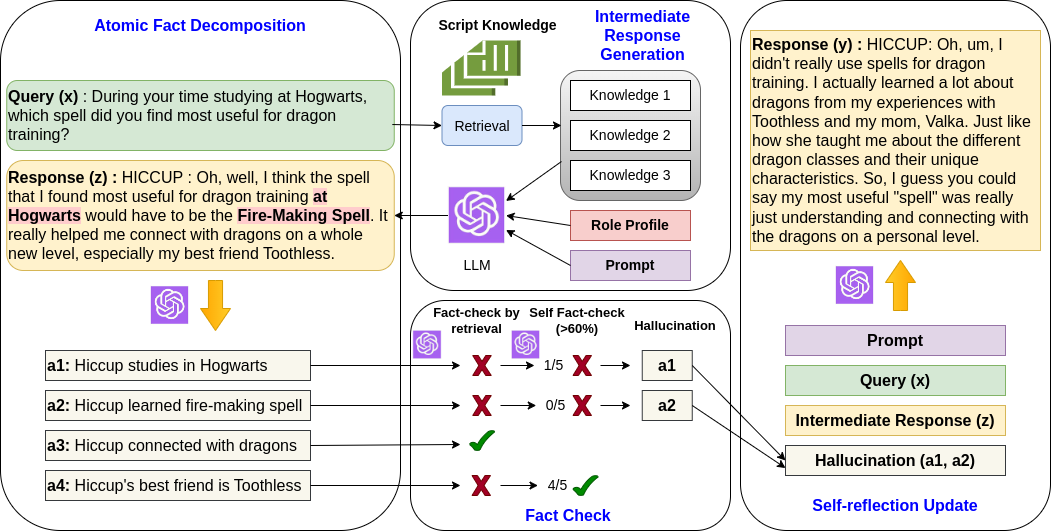}
	\caption{An overview of \proposed, which performs parametric and non-parametric verification of atomic facts.}
	\label{fig:proposed_rolefact}
 \vspace{-1em}
\end{figure*}
\proposed aims to mitigate hallucination by modulating the influence of parametric knowledge within character responses. The overview of the method is shown in Figure~\ref{fig:proposed_rolefact}. \proposed generates a response using a character profile and retrieved knowledge and updates the response based on atomic fact verification using both parametric and non-parametric retrieved knowledge. All facts supported by retrieved knowledge may remain in the final response. Facts that are only supported by parametric knowledge of LLMs remain in the final response if the confidence is above a calibrated threshold. 

Let $x\in X$ represent a query and $y\in Y$ a response generated by the character. Let  $\mathit{RET}: X\times \mathbb{N}\times U\rightarrow U_n$ serve as a retrieval function that fetches the most relevant $n\in \mathbb{N} $ documents relevant to the query $x$ from a set of non-parametric knowledge base $U$. The retrieved knowledge may be denoted as $u_n\in U_n$ where $U_n\subset U$. Let $p \in P$ denote the role profile description and $z\in Z$ an intermediate response from the character. The response generation function $\mathit{IRG}: X\times P\times U_n\rightarrow Z$ utilizes the query $x$, the role profile $p$, and the retrieved knowledge $u_n$ to produce an intermediate response $z$. The prompt used for $\mathit{IRG}$ is shown below (ignoring line breaks).

\vspace{0.4em}
\fbox{\begin{minipage}{0.92\columnwidth}
$\mathit{IRG}$: You will role-play as <role\_name> from <story\_title>. Your task is to respond to the following dialogue context. If the question matches a scene from your storyline, please reuse the original lines from the story. You will respond and answer like <role\_name> using the tone, manner, and vocabulary <role\_name> would use. Your character description is: <$p$>. Relevant scenes for the given context are as follows: <$u_n$>. Dialogue context: <$x$>.
\end{minipage}}
\vspace{0.2em}

The response $z$ is decomposed into a list of atomic facts $A_z$ with a function $\mathit{DEC}: Z \rightarrow A_z$. 

\vspace{0.4em}
\fbox{\begin{minipage}{0.92\columnwidth}
$\mathit{DEC}$: I will give you an utterance from a movie or a play. Your task is to provide me with a list of atomic facts expressed in the given utterance. Each atomic fact should be described in a name-only third-person format. Utterance: <$z$>.
\end{minipage}}

Each atomic fact $a \in A_z$ is verified against the retrieved non-parametric knowledge $U_n$ and parametric knowledge $V$ associated with an LLM. The functions for fact-checking against non-parametric and parametric knowledge bases are denoted as $\mathit{FCR}: A_z\times U_n\rightarrow \{0,1\}$ and $\mathit{FCS}: A_z\times V\rightarrow \{0,1\}$ respectively. A set of verified atomic facts, $A_y$, is constructed such that if $a$ is supported by $U_n$, it is added to $A_y$. Conversely, if $a$ is not supported by $U_n$, it is only included in $A_y$ if it appears to be supported by $V$ within a reasonable confidence threshold $t$. Let's assume the function $\mathit{FCS}$ is run $m$ times using an LLM and $a$ is supported $k$ times by $V$. The fact $a$ is added to $A_y$ if $\frac{k}{m}\geq t$. The sample size $m$ and the confidence threshold $t$ are calibrated on a validation set.

\vspace{0.4em}
\fbox{\begin{minipage}{0.92\columnwidth}
$\mathit{FCR}$: Consider the given statement and the evidence knowledge sources. Indicate whether the statement is supported by the knowledge sources. Negation of a false statement should be considered supported. Statement: <$a$>. Evidence Knowledge <$u_n$>.
\end{minipage}}

\fbox{\begin{minipage}{0.92\columnwidth}
$\mathit{FCS}$: Consider the given statement by <role\_name> from <story\_title>. Indicate whether the statement is supported by the storyline of <story\_title>. Negation of a false statement should be considered supported. Statement: <$a$>.
\end{minipage}}
\vspace{0.2em}

The final response $y$ is generated by updating the intermediate response $z$ via self-reflection conditioned on $x$. During self-reflection, all unverified atomic facts $A_z \setminus A_y$ are removed from $z$. The self-reflection update function is denoted as $\mathit{SRU}: X\times Z\times (A_z\setminus A_y) \rightarrow Y$.

\vspace{0.4em}
\fbox{\begin{minipage}{0.92\columnwidth}
$\mathit{SRU}$: Consider the following response generated by an AI role-playing as <role\_name> from <story\_title>. The response may contain one or more unsupported claims as a result of hallucination. The unsupported claims are listed below. Rewrite the response to remove all the unsupported claims from the response. If the hallucination stems from a wrong assertion made in the original query, feel free to clarify that. Original query <$x$>. Response <$z$>. Unsupported Claims <$A_z\setminus A_y$>.  
\end{minipage}}
\vspace{0.2em}

\begin{algorithm}
\caption{\proposed Algorithm}
\begin{algorithmic}[1]
\Require $x, p, n, m, t $
\Statex $x \in X \gets $ Input query
\Statex $p \in P \gets $ Role profile
\Statex $n \in \mathbb{N} \gets $ Number of documents
\Statex $u_n \gets $ Retrieved documents
\Statex $z \in Z \gets $ Intermediate response
\Statex $m \gets $ Sample size
\Statex $t \gets $ Confidence threshold

\State $u_n \gets \Call{ret}{x,n}$\Comment{perform retrieval}
\State $z \gets \Call{irg}{x,p,u_n}$\Comment{intermediate response}
\State $A_z \gets \Call{dec}{z}$\Comment{decompose facts}
\State $A_y \gets \{\}$

\For{$a \in A_z$}
    \State $s_1 \gets \Call{fcr}{a,u_n}$ \Comment{retrieval fact-check}
    \If{$s_1 = 1$}
        \State $A_y \gets A_y \cup \{a\}$ \Comment{add supported fact}
        \State \textbf{continue}
    \EndIf
    \State $k \gets 0$
    \For{$i = 1$ \textbf{to} $m$}
        \State $s_2 \gets \Call{fcs}{a,V}$ \Comment{self fact-check}
        \State $k \gets k + s_2$
    \EndFor
    \If{$\frac{k}{m} \geq t$}
        \State $A_y \gets A_y \cup \{a\}$ \Comment{add supported fact}
    \EndIf    
\EndFor
\State $y \gets \Call{sru}{x,z,A_z\setminus A_y}$  \Comment{update response}
\State \textbf{return} $y$
\end{algorithmic}
\end{algorithm}
\section{The SGR Dataset}
The summary of the SGR dataset is shown in Table~\ref{tab:sgr_stats}. The scripts for the dataset are collected from IMSDb~\cite{imsdb}, Screenplay~\cite{screenplay}, and Open Source Shakespeare~\cite{oss2024}. PDF scripts are converted to text with gpt-4-vision-preview~\cite{gpt4vision}. Scripts are manually grouped into storylines. For example, eight scripts for `Harry Potter' movies form one storyline. There are 1152 unique stories in the dataset. Scripts are split into scenes and then scenes into knowledge events. A knowledge event can be either a speech event or a non-speech event. The relevant character is identified for each speech event. Script segmentation and character identification are performed with gpt-3.5-turbo-0125~\footnote{We denote this model as GPT-3.5 for the rest of the paper}~\cite{gpt-models}. The dataset contains 2.4 million knowledge events of which 1.1 million are speech events. Time annotations are added to each event, starting at zero at the beginning of each story.

\begin{table}
\resizebox{\columnwidth}{!}{
\begin{tabular}{lrlr}
\toprule
Attribute  & Statistics & Attribute  & Statistics \\ 
\midrule
Knowledge  & 2.4 M & Story-line  & 1,152 \\ 
Speech     & 1.1 M & Characters & 2,000 \\
Non speech & 1.3 M & Interviews & 72,000  \\
\bottomrule
\end{tabular}
}
\caption{SGR Dataset}
\vspace{-1em}
\label{tab:sgr_stats}
\end{table}

The dataset has four tasks, namely 1) adversarial interview (ADV) 2) open-ended interview (OEI) 3) dialogue completion (DC) 4) scene-grounded interview (SGI). The adversarial task tries to induce cross-universe hallucination while the open-ended task ensures the system does not become too conservative due to hallucination mitigation. The dialogue completion and scene-grounded tasks try to induce temporal hallucination. The adversarial and open-ended interview questions are constructed by prompting GPT-4~\cite{gpt-models} in a script-agnostic fashion. Dialogue completion task prompts the characters to respond to dialogues extracted from the script. Scene-grounded interview questions are constructed by prompting GPT-4 with specific scenes. The questions aim to make characters answer questions about a specific scene in the story. For adversarial and open-ended tasks, we assume that characters have access to all knowledge in the story. For dialogue completion and scene-grounded interviews, we assume characters only have knowledge of events leading up to the current scene. We have 18,000 samples for each task type with a total of 72,000 interviews. Examples of all tasks with relevant character responses can be found in Appendix~\ref{sec:appendix}.
\begin{table*}
\resizebox{0.95\textwidth}{!}{
\begin{tabular}{llcccccccc}
\toprule
 &  & \multicolumn{2}{c|}{Adversarial} & \multicolumn{2}{c|}{Open Ended} & \multicolumn{2}{c|}{Dialogue Completion} & \multicolumn{2}{c}{Scene Grounded} \\
 \midrule
LLM & Method & Fact Score & SFPR & Fact Score & SFPR & Fact Score & SFPR & Fact Score & SFPR \\
\midrule
\multirow{4}{*}{Vicuna-7b} & Baseline & 0.41 & 3.7 & 0.72 & 6.5 & 0.52 & \textbf{6.4} & 0.65 & 5.2 \\
 & + KGR & 0.45 & 3.3 & 0.74 & 6.6 & 0.54 & 5.7 & 0.68 & 4.5 \\
 & + SR & 0.44 & 3.5 & 0.70 & 6.7 & 0.51 & 6.2 & 0.62 & 5.1 \\
 & RoleFact & \textbf{0.49} & \textbf{3.8} & \textbf{0.81} & \textbf{6.8} & \textbf{0.63} & 6.3 & \textbf{0.77} & \textbf{5.4} \\
 \midrule
\multirow{4}{*}{Llama-3-8b} & Baseline & 0.50 & \textbf{5.0} & 0.70 & 7.7 & 0.52 & 3.4 & 0.62 & \textbf{6.0} \\
 & + KGR & 0.56 & 4.1 & 0.72 & 7.4 & 0.50 & 3.2 & 0.65 & 5.1 \\
 & + SR & 0.54 & 4.3 & 0.71 & 7.5 & 0.51 & 3.3 & 0.64 & 5.6 \\
 & RoleFact & \textbf{0.62} & 4.8 & \textbf{0.79} & \textbf{7.8} & \textbf{0.57} & \textbf{3.5} & \textbf{0.71} & 5.9 \\
 \midrule
\multirow{4}{*}{GPT-3.5} & Baseline & 0.61 & \textbf{3.8} & 0.76 & 5.8 & 0.65 & \textbf{4.2} & 0.74 & \textbf{4.8} \\
 & + KGR & 0.65 & 2.8 & 0.79 & 5.6 & 0.68 & 3.8 & 0.77 & 4.3 \\
 & + SR & 0.63 & 3.1 & 0.78 & 5.7 & 0.66 & 3.9 & 0.75 & 4.3 \\
 & \proposed & \textbf{0.72} & 3.7 & \textbf{0.88} & \textbf{5.8} & \textbf{0.77} & 4.0 & \textbf{0.85} & 4.6\\ 
 \bottomrule
\end{tabular}
}
\caption{Factual precision (Fact Score ↑) and informativeness (SFPR ↑) of \proposed on all tasks in the SGR dataset. All results here are produced with BM25 retrieval.}
\vspace{-1em}
\label{tab:rolefact_performance}
\end{table*}

\section{Experiments}

We conduct experiments with three LLM backbones, namely \textbf{Vicuna-7B-1.5}~\cite{zheng2023judging}, \textbf{Llama-3-8B-Instruct}~\cite{meta-llama3}, and \textbf{GPT-3.5-Turbo}. For retrieval, we consider \textbf{BM25}~\cite{robertson2009probabilistic}, \textbf{S-BERT}~\cite{reimers2019sentence}, and \textbf{Contriever}~\cite{izacard2022unsupervised}. 1,000 randomly chosen samples from each task type constitute the validation set. The validation set is used for calibrating the confidence threshold, sample size, the retrieval method, and the number of documents to retrieve. Decoding for character response generation is performed with temperature = 0.7 and top\_p = 0.95 for all models.

\paragraph{Baselines} Our primary baseline is similar to \citet{li2023chatharuhi}. We prompt an LLM with a role prompt, a role profile, and retrieved dialogues relevant to the query. We also consider two additional variants of the primary baseline designed to improve the factual precision of role-play. The first variant rewrites the response of the primary baseline by removing facts absent from the retrieved knowledge. This is denoted as knowledge-guided rewriting \textbf{(Baseline + KGR)} baseline. The second variant updates the primary response using self-reflection to identify and mitigate hallucination based on LLM self-confidence. This is denoted as the self-reflection \textbf{(Baseline + SR)} baseline.

\paragraph{Evaluation Metrics} We measure factual precision with \textbf{Fact Score}~\cite{min2023factscore}. All atomic facts are converted to a `name-only third person' format for fact-checking. We calculate the number of supported atomic facts per response \textbf{(SFPR)} to evaluate informativeness. We also measure temporal hallucination rate \textbf{(THR)}, which is defined as the number of atomic facts associated with temporal hallucination for every 100 responses. THR was determined manually after analyzing unsupported facts from 500 samples from dialogue completion and scene-grounded tasks. For human evaluation, we use a rating between one and seven to rate 200 responses in terms of factuality, informativeness, and speaker style imitation.

\paragraph{Results} Table~\ref{tab:rolefact_performance} shows the performance of \proposed on the SGR dataset. \proposed outperforms all three baselines in factual precision while remaining competitive with the primary baseline in terms of informativeness. For GPT-3.5, the relative improvement of \proposed over the primary baseline in 18.0\%, 15.7\%, 18.4\%, and 14.8\% for adversarial, open-ended, dialogue completion, and scene-grounded tasks respectively. The second best approach in terms of factuality is KGR. However, KGR is also the least informative since it only allows facts supported by retrieved knowledge. Comparing across LLMs, we find that GPT-3.5 is the best for factuality while Llama-3-8B is more informative (30\% more SFPR for adversarial tasks).

\begin{table}
\begin{center}
\resizebox{0.7\columnwidth}{!}{
\begin{tabular}{llrr}
\toprule
LLM & Method & SGI & DC \\ 
\midrule
\multirow{2}{*}{Vicuna-7b} & Baseline & 39.2 & 59.8 \\
 & RoleFact & \textbf{33.4} &\textbf{52.0} \\
 \midrule
\multirow{2}{*}{Llama-3-8b} & Baseline & 29.6 & 65.3 \\
 & RoleFact & \textbf{17.1} & \textbf{41.7} \\
 \midrule
\multirow{2}{*}{GPT-3.5} & Baseline & 26.5 & 57.2 \\
 & RoleFact & \textbf{14.7} & \textbf{38.5} \\
 \bottomrule
\end{tabular}  
}
\end{center}
\caption{Temporal hallucination rate (THR ↓) on scene-grounded interview and dialogue completion tasks.}
\vspace{-1em}
\label{tab:temporal_hallucination}
\end{table}

\begin{figure}
\centering
\includegraphics[width=0.95\columnwidth]{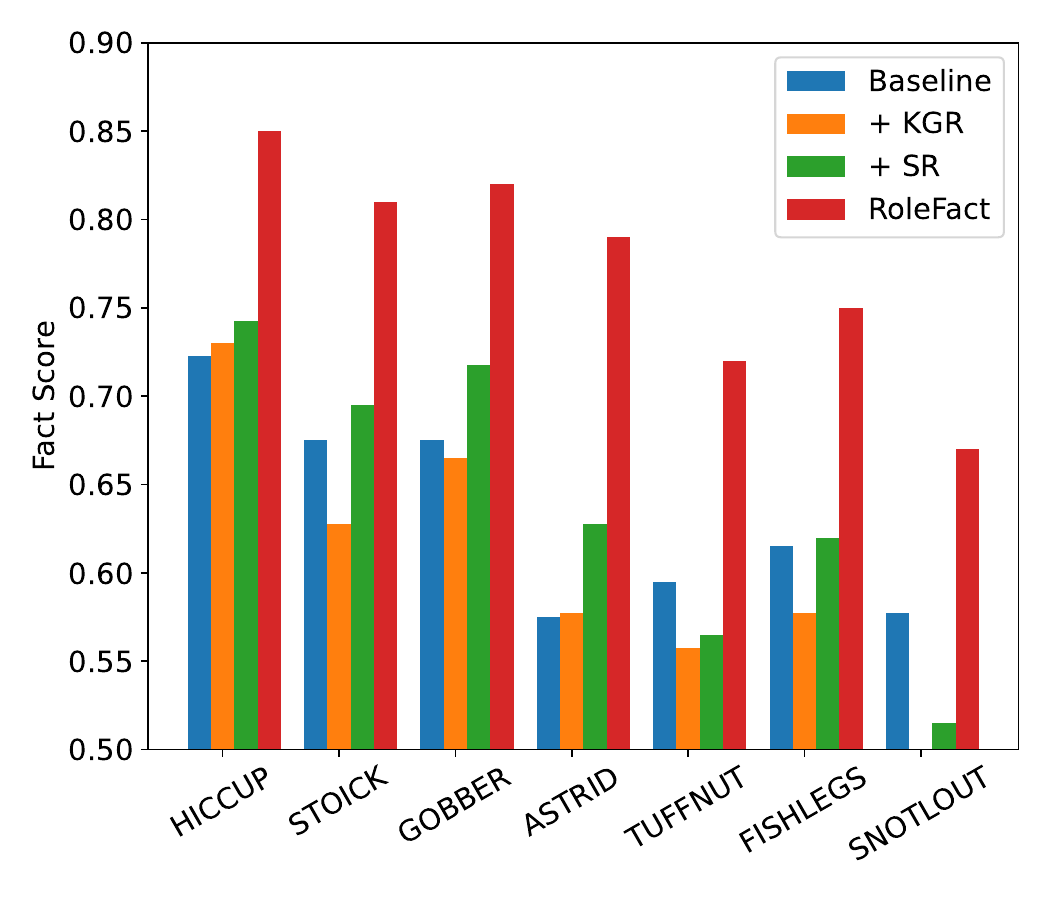}
\caption{Performance by role popularity (decreasing popularity left to right, adversarial task, GPT-3.5).}
\vspace{-1em}
\label{fig:long_tail_proposed}
\end{figure}

\proposed decreases temporal hallucination on dialogue completion and scene-grounded tasks, as shown in Table~\ref{tab:temporal_hallucination}. For GPT-3.5, the relative reduction of hallucination is 32.7\% and 44.5\% for dialogue completion and the scene-grounded setting respectively.
\proposed improves performance for less popular characters (example shown in Figure~\ref{fig:long_tail_proposed}). Excluding the most popular ten characters per story, the relative improvement is 22.9\%.

\begin{figure*}
\centering
\includegraphics[width=\textwidth]{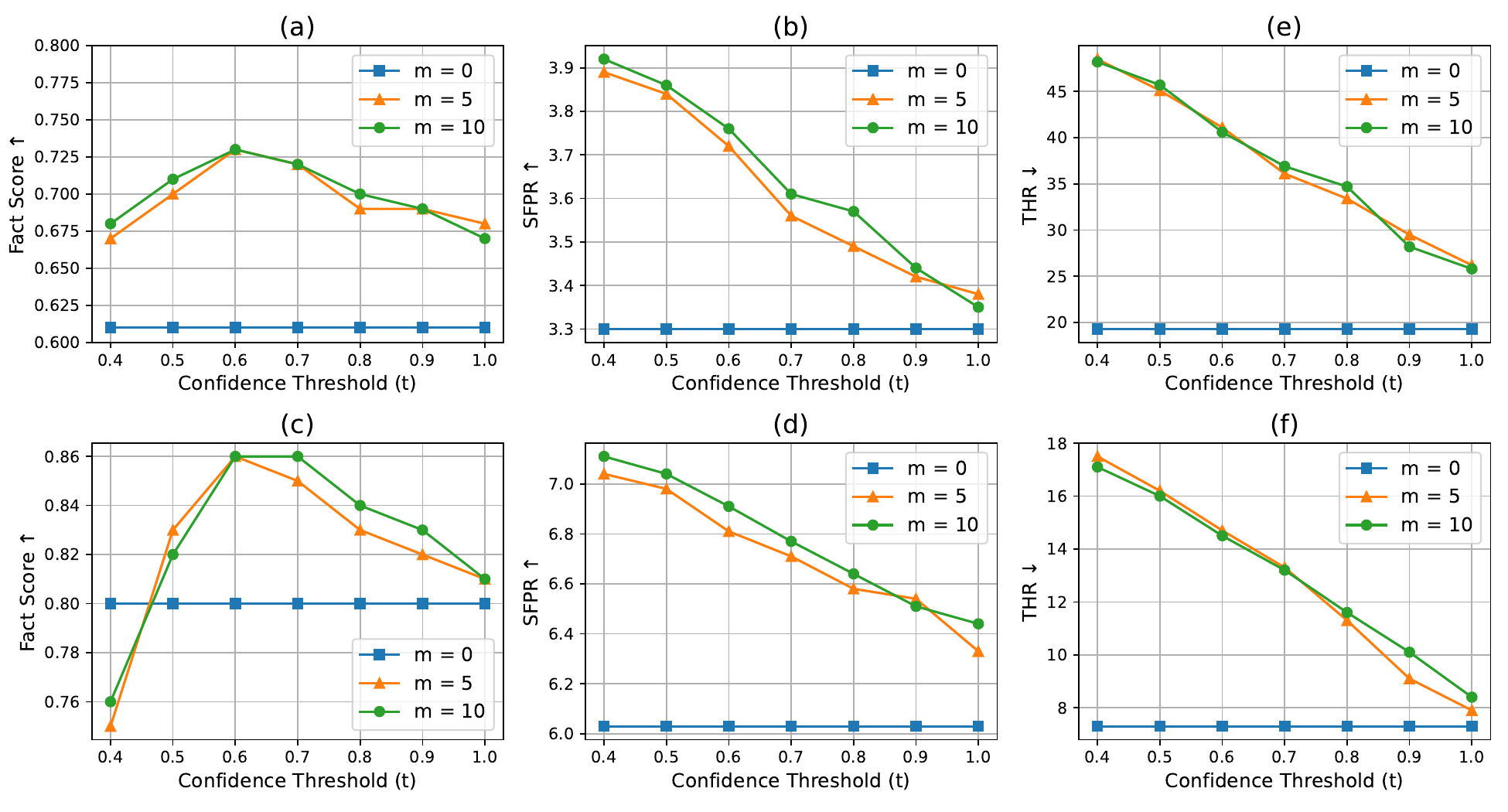}
\caption{Calibration for confidence threshold $(t)$ and sample size $(m)$ on the validation set with GPT-3.5. $(a,b)$ shows fact score and SFPR on adversarial task while $(c,d)$ shows the same on open ended task. The temporal hallucination rates for dialogue completion and scene-grounded tasks are shown by $e$ and $f$ respectively.}
\label{fig:confidence_sample_size}
\vspace{-1em}
\end{figure*}

\paragraph{Hyper-parameter Tuning} Figure~\ref{fig:confidence_sample_size} shows the impact of confidence threshold $(t)$ and sample size $(m)$ on factual precision, informativeness, and temporal hallucination. Setting $m=0$ indicates all facts must be strictly supported by retrieved knowledge (similar to KGR). For $m>0$, a small $t$ increases the influence of parametric knowledge. This benefits informativeness but has a higher temporal hallucination. Conversely, a high $t$ decreases the influence of parametric knowledge. This means less temporal hallucination but also less informative character responses.  We choose $t=0.6$ since it has the best fact score for adversarial and open-ended tasks while finding a middle ground between informativeness and temporal hallucination.

\begin{figure}
\centering
\includegraphics[width=0.9\columnwidth]{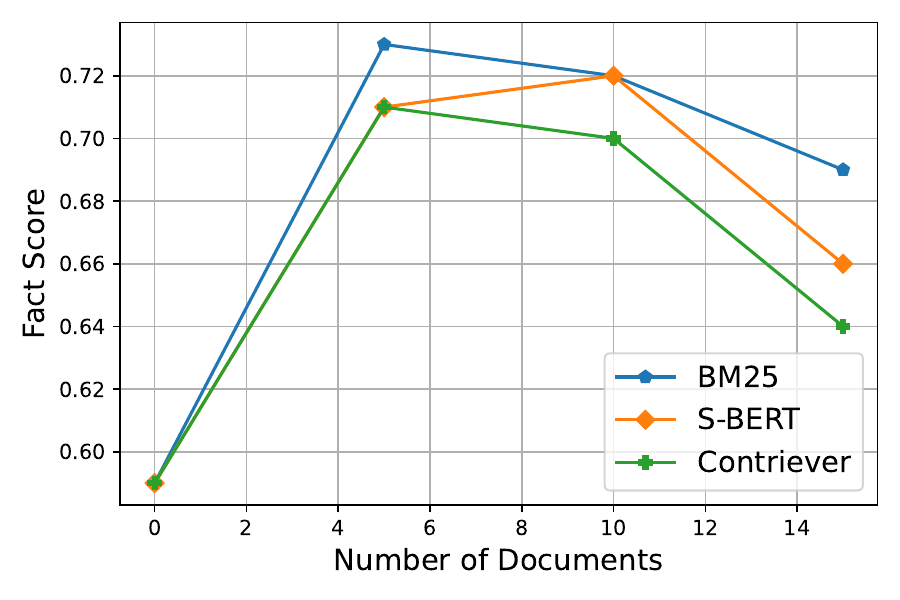}
\caption{Performance of retrieval methods w.r.t number of retrieved documents (adversarial task, GPT-3.5).}
\label{fig:retrieval_performance}
\end{figure}

Figure~\ref{fig:retrieval_performance} shows the performance of \proposed with respect to different retrieval approaches and number of retrieved documents. BM25 outperforms off-the-shelf dense retrieval methods in most cases. Interestingly, we find that increasing the number of retrieved documents beyond five may hurt factual precision. This phenomenon may be explained by similar findings for RAG~\cite{shuster2021retrieval} where increased irrelevant context added to the prompt exacerbates hallucination. 

\paragraph{Ablation study} We perform an ablation study to identify the contribution of parametric knowledge, non-parametric retrieved knowledge, and role profile to the supported facts in the \proposed responses. This is shown in Table~\ref{tab:ablation}. The most significant performance drop happens when we make the prompts anonymous (0.72 $\rightarrow$ 0.56), indicating that the largest share of facts may be attributed to parametric knowledge. The fact score drops to 0.58 when retrieved knowledge is removed from the prompt, indicating that retrieved knowledge is almost equally important. The role profile has the lowest impact on performance (0.72 $\rightarrow$ 0.64).

\begin{table}
\begin{center}
\resizebox{0.65\columnwidth}{!}{
\begin{tabular}{@{}lll@{}}
\toprule
Method & Fact Score & SFPR \\ \midrule
RoleFact & 0.72 & 3.7 \\
+ anonymous & 0.56 & 3.3 \\ 
- retrieval & 0.58 & 3.2 \\
- role profile & 0.64 & 2.9 \\
\bottomrule
\end{tabular}
}
\end{center}
\caption{Ablation study (adversarial task, GPT-3.5)}
\vspace{-0.5em}
\label{tab:ablation}
\end{table}

\paragraph{Human Evaluation} Human ratings on a scale of one to seven in terms of factuality, informativeness, and speaker style similarity are given in Table~\ref{tab:human}. \proposed outperforms all three baselines in terms of factuality with a score of 6.1, compared to 4.9, 6.0, and 5.6 scores from the baselines. While the KGR baseline also achieves a competitive rating, the responses are less informative and demonstrate poor speaker style when retrieved knowledge does not contain the required information.

\begin{table}
\begin{center}
\resizebox{\columnwidth}{!}{
\begin{tabular}{lccc}
\toprule
Method & Factuality & Informativeness & Speaker Style \\ \midrule
Baseline & 4.9 & \textbf{5.8} & 5.2 \\
+ KGR & 6.0 & 5.2 & 4.3\\ 
+ SR & 5.6 & 5.4 & 5.1\\
\proposed & \textbf{6.1} & 5.7 & \textbf{5.3}\\
\bottomrule
\end{tabular}
}
\end{center}
\caption{Human ratings on a scale of one to seven}
\vspace{-0.5em}
\label{tab:human}
\end{table}

\paragraph{Case Study} Case study in shown in Table~\ref{tab:case_study0}. \textbf{Case 1} is an adversarial question where Anakin from `Star Wars' is asked how his friendship with Spock from `Star Trek' influenced his decisions during the events of Clone Wars. The baseline response suffers from cross-universe hallucination and mistakenly acknowledges the friendship. \proposed not only denies the wrong assertion made in the question but also clarifies that his decisions during the Clone Wars were actually influenced by training from his mentor Obi-Wan and other Jedi. \textbf{Case 2} is a dialogue completion task taken from a dinner scene of the `How to Train Your Dragon' movie, taking place after Hiccup's very first day of dragon training. In the baseline response, the target character Ruffnut refers to Hiccup as their fearless leader. This is an example of temporal hallucination induced by parametric memory of LLMs since the knowledge of Hiccup's fearlessness and leadership is not made available to the LLM via retrieval. Unlike the baseline, \proposed is able to avoid mentioning future information. \textbf{Case 3} shows a scene grounded interview from the `Harry Potter' series. Shortly after being sorted into his house, Harry is asked how he perceives Snape's strange look towards him. The baseline response reveals information about future events that was not made available to it via retrieval such as Harry's connection with Voldemort, Snape's final moment, etc. \proposed speculates that the look may be associated with dislike or suspicion which is much more aligned with Harry's understanding of the situation at the time. More examples including the responses from the other two baselines can be found in Appendix~\ref{sec:appendix}.

\begin{table*}
\resizebox{0.95\textwidth}{!}{
\begin{tabular}{p{5cm}p{5cm}p{5cm}p{5cm}}
\toprule
\multicolumn{4}{p{21cm}}{\textbf{[Case 1: Adversarial Interview] \textcolor{blue}{Question: In what ways did your friendship with Spock influence your decisions during the Clone Wars?}} (\textbf{Target Character:} ANAKIN from `Star Wars')} \\
\midrule
 \textbf{Baseline} & \textbf{Baseline + KGR} & \textbf{Baseline + SR} & \textbf{RoleFact} \\
 \bluebold{ANAKIN}: \redul{My friendship with} \redul{Spock was a powerful influence} \redul{on me. He was a wise mentor} \redul{and a trusted ally}, guiding me through the complexities of war and teaching me the value of loyalty and ... \redul{His logical} \redul{approach to situations} balanced out my impulsiveness, allowing us to make strategic decisions that benefited our cause ... & \bluebold{ANAKIN}: My apologies, but I do not have any recollection of a friendship with Spock during the Clone Wars. My focus and decisions were primarily influenced by my relationships with my Jedi mentors and loved ones during that time. & \bluebold{ANAKIN}: I'm sorry, but I do not have information about a character named Spock or any influence he may have had during the Clone Wars. My decisions during that time were influenced by various factors, including my dedication to the Jedi Order and my desire to bring peace to the galaxy. & \bluebold{ANAKIN}: I'm sorry, but I believe you are confused. I am Anakin Skywalker, not involved with Spock in any way. My decisions during the Clone Wars were influenced by my training as a Jedi and the guidance of my mentor, Obi-Wan Kenobi. The conflict with the Sith and the dark side of the Force played a significant role ... \\
 \midrule
 \multicolumn{4}{p{21cm}}{\textbf{[Case 2: Dialogue Completion] \textcolor{blue}{Question: GOBBER: The dragon manual. Everything we know about every dragon we know of ... TUFFNUT: (you've got to be kidding) Wait, you mean read?}} ( \textbf{Target Character:} RUFFNUT from `How to Train Your Dragon' after first day of dragon training)} \\
\midrule
\textbf{Baseline} & \textbf{Baseline + KGR} & \textbf{Baseline + SR} & \textbf{RoleFact} \\
 \bluebold{RUFFNUT}: Reading? Ugh, that's for the weaklings. I prefer to learn on the fly, you know, with a little hands-on experience. \redul{Plus, who needs a dragon manual} \redul{when we've got our own fearless} \redul{leader, Hiccup, to show us the} \redul{ropes?} ... &  \bluebold{RUFFNUT}: Ugh, reading? That's for the weak and boring. I'd rather be out there causing some chaos with the dragons. Who needs a stupid manual when you can figure things out on the fly? ... &  \bluebold{RUFFNUT}: Ugh, reading? Seriously? Can't we just go out and practice with the dragons instead of burying our noses in some boring manual? Let's go have some real fun, like... causing chaos with the twin ... &  \bluebold{RUFFNUT}: Ugh, reading? Seriously? Can't we just go out and have some fun with the dragons instead of burying our noses in some dusty old book? Who needs to study when you can just learn on the fly ... *rolls eyes* \\
 \midrule
 \multicolumn{4}{p{21cm}}{\textbf{[Case 3: Scene Grounded Interview] \textcolor{blue}{Question: What do you believe Snape's look towards you signifies?}} ( \textbf{Target Character:} HARRY from `Harry Potter', during the sorting hat ceremony in the first year)} \\
\midrule
\textbf{Baseline} & \textbf{Baseline + KGR} & \textbf{Baseline + SR} & \textbf{RoleFact} \\
 \bluebold{HARRY}: \redul{I believe Snape's look} \redul{towards me signifies a mixture} \redul{of guilt, duty, and perhaps even} \redul{a hint of remorse. He knew the} \redul{truth about the prophecy and the} \redul{connection between Voldemort} \redul{and myself}, and he carried the burden of that knowledge heavily ... And in the end, \redul{his final} \redul{actions spoke volumes} about the complexity of his character and his true loyalty. &  \bluebold{HARRY}: \redul{Snape's look towards} \redul{me signifies a complicated mix} \redul{of emotions and history between} \redul{us. He has always had a tumult-} \redul{uous relationship with me, given} \redul{his past as a Death Eater and his} \redul{dual allegiance to both Dumble-} \redul{dore and Voldemort.} I believe his look towards me carries a sense of duty, guilt, and perhaps even a glimmer of remorse for his actions ... &  \bluebold{HARRY}: \redul{I believe Snape's look} \redul{towards me signifies a complex} \redul{mixture of emotions. He harbors} \redul{a deep sense of guilt and remorse} \redul{for his past actions, especially} \redul{towards my parents, but also a} \redul{sense of duty and loyalty towards} \redul{Dumbledore's cause. There is} \redul{also an underlying sense of} \redul{protectiveness and care that he} \redul{feels towards me, despite his} \redul{outward demeanor ...} &  \bluebold{HARRY}: Well, I think Snape's look towards me signifies a great deal of dislike and suspicion. \\
 \midrule
  \multicolumn{4}{p{21cm}}{\textbf{[Case 4: Open Ended Interview] \textcolor{blue}{Question: Hagrid, what is your fondest memory with Harry, Ron, and Hermione?}} (\textbf{Target Character:} HAGRID from `Harry Potter')} \\
\midrule
\textbf{Baseline} & \textbf{Baseline + KGR} & \textbf{Baseline + SR} & \textbf{RoleFact} \\
 \bluebold{HAGRID}: Oh, well, there's so many memories with those three, but I reckon one of my fondest is when we rescued Norbert together. Harry, Ron, and Hermione were so brave and resourceful. It was a real adventure, that was. \redul{And I'll always remember} \redul{the look on their faces when we} \redul{were flying on Norbert's back.} Good times, they were. &  \bluebold{HAGRID}: ... It was when Buckbeak was in grave danger of being executed unjustly. They stood by me, showed compassion for creatures like Buckbeak, and even tried to comfort me in their own way. It meant a lot to me, seein' their loyalty and compassion. &  \bluebold{HAGRID}: Oh, bless 'em, so many memories I have with Harry, Ron, and Hermione. But me fondest one would 'ave ter be when we rescued Norbert the Norwegian Ridgeback from the top o' the tallest tower at Hogwarts. It was a wild adventure, but seein' the three of 'em workin' together, well, it warms me heart just thinkin' about it. &  \bluebold{HAGRID}: I reckon it would have to be when we all snuck down ter visit Buckbeak before his trial. 'Twas a risky move, but seein' 'em standin' up for a creature like that, showin' true courage and loyalty, it truly touched me heart ... Aye, that moment will always be close ter me heart ... \\
\midrule
\multicolumn{4}{p{21cm}}{\textbf{[Case 5: Adversarial Interview] \textcolor{blue}{Question: Please describe the wedding ceremony between you and Queen Elsa from Arendelle. How did you two meet?}} (\textbf{Target Character:} GOBBER from `How to Train Your Dragon')} \\
\midrule
 \textbf{Baseline} & \textbf{Baseline + KGR} & \textbf{Baseline + SR} & \textbf{RoleFact} \\
 \bluebold{GOBBER}: ... Well now, that's a tale for the ages! \redul{We met on a} \redul{chilly winter day} .... \redul{Queen Elsa} \redul{and I bonded over our shared} \redul{love for mystical creatures... we} \redul{found ourselves falling in love.} \redul{The wedding ceremony was a} \redul{grand affair, with dragons soaring} \redul{through the sky and ...} Aye, it was a day to remember, it was. &  \bluebold{GOBBER}: Oh, I'm afraid you've got me confused with someone else, laddie! I haven't tied the knot with Queen Elsa from Arendelle. But I can tell you a thing or two about Viking weddings if you're interested! &  \bluebold{GOBBER}: Wedding ceremony between me and Queen Elsa? Now, lad, I'm afraid you've got the wrong tale there. I'm from Berk, not Arendelle, and I've never crossed paths with Queen Elsa. As for me and weddings, well, let's just say my adventures with dragons have kept me plenty occupied. &  \bluebold{GOBBER}: Oh, laddie, ye got me there! I'm afraid I have no knowledge o' any wedding ceremony with Queen Elsa from Arendelle. I reckon ye might be mixin' me up with someone else from a different tale. But if ye have any questions about dragons or blacksmithin', I'd be more than happy to chat about that! \\
 \bottomrule
\end{tabular}
}
\caption{Case Study, hallucinations are in red and underlined.}
\label{tab:case_study0}
\end{table*}
\section{Related Work}

\paragraph{Language Models for Role-play}
Building autonomous agents that can simulate human behavior with individualized traits is a long-standing challenge~\cite{Minsky1986TheSO}.
Prior works in NLP build conversational agents grounded with various individualized traits, such as background stories~\cite{zhang-etal-2018-personalizing, park2023generativeagents}, demographic information~\cite{hwang-etal-2023-aligning,chuang2024beyond}, social roles~\cite{zhou2024sotopia}, and response style constraints~\cite{han-etal-2022-meet}. In this work, we focus on building dialogue systems that are aligned with user perceptions of fictional characters, known as character role-play~\cite{shao2023characterllm, chen-etal-2023-large, li2023chatharuhi}. In terms of methodology, prior works can be categorized as training-free role-prompting~\cite{wang2024rolellm,zhou2023characterglm,li2023chatharuhi} or instruction-tuning~\cite{shao2023characterllm,lu-etal-2024-large}. Role-prompting includes static character descriptions~\cite{zhou2023characterglm,wang2024rolellm} and retrieved past experiences~\cite{li2023chatharuhi,wang2024rolellm}. In terms of role-playing objectives, prior works focused on personality traits~\cite{sang-etal-2022-mbti,serapio2023personality}, speaker style~\cite{chen-etal-2023-large,zhao-etal-2024-narrativeplay}, and character fidelity~\cite{wang-etal-2024-incharacter,shao2023characterllm}. We specifically focus on character fidelity with respect to story-specific knowledge and relevant timelines. A character role-play dialogue system should produce responses aligned with facts related to its assigned character and storyline and it should demonstrate a lack of knowledge about things that do not apply to them.~\cite{shao2023characterllm}. However, most existing methods of role-play lack mechanisms to reduce parametric memory-induced hallucination. Moreover, heavy reliance on parametric memory leads to poor performance for less popular characters since LLMs may not have adequate parametric knowledge about them. To the best of our knowledge, our work is the first one to modulate the influence of parametric knowledge to mitigate hallucination. This is also the first dataset to facilitate role-playing for less popular characters. 

\paragraph{Factuality in Natural Language Generation}
Beyond conversational systems, improving factuality in generated contents is an open problem in NLP, benefiting domains such as summarization~\cite{cao2018faithfultooriginal, dong-etal-2020-multi}, knowledge grounded dialogue systems~\cite{dziri-etal-2021-neural}, and question answering~\cite{honovich-etal-2021-q2, gao2024spuq}. While fine-grained evaluation of factuality based on atomic facts has been studied in NLP~\cite{min2023factscore}, there is currently no dataset for character role-play that facilitates similar evaluation of factuality. Therefore, prior works on role-play~\cite{wang2024rolellm,li2023chatharuhi,shao2023characterllm} have relied on rating-based evaluation of factuality which is expensive for humans. Our dataset is the first one to facilitate automated fine-grained analysis of factuality by matching decomposed atomic facts against script-specific knowledge. 

\paragraph{Retrieval Augmented Methods}
Our work is also related to Retrieval-augmented methods in NLP that improve model output by incorporating retrieved external knowledge, typically for generation tasks~\cite{Khandelwal2020Generalization, lewis2020retrieval, Guu2020Realm}.
While this class of method has been used in building role-playing agents~\cite{zhang-etal-2018-personalizing, park2023generativeagents, shao2023characterllm, chen-etal-2023-large, li2023chatharuhi}, these works perform retrieval in a time-agnostic manner. In order to study the temporal dimension of character role-play (i.e. character development), we need a dataset where script knowledge is aligned to temporal annotations. Our dataset provides temporal annotations with utterance-level granularity. Contemporary to our work, \citet{ahn2024timechara} also release a dataset for time-sensitive role-play. However, their dataset only covers 14 characters and does not contain utterance or scene-level granularity of temporal annotations.  
\section{Conclusion}
In this work, we focus on the mitigation of hallucination for character role-play. We introduce the SGR dataset which provides story-specific knowledge aligned with task-specific interviews for a diverse set of characters. The dataset is specifically designed for studying hallucinations of various types such as cross-universe hallucinations and temporal hallucinations. SGR is the first dataset to facilitate automated evaluation of hallucination and analysis of factuality for less popular characters. We propose \proposed, a role-playing method that mitigates hallucination via parametric and non-parametric verification of atomic facts. \proposed improves the overall factual precision by up to 18.4\%. An interesting direction for future research is to instruction-tune an open-weight LLM with script-specific knowledge and task-specific interviews from the SGR dataset for generating factual and informative responses.

\section{Limitations}
We find that the factual precision of \proposed is sensitive to retrieval quality, as shown in Figure~\ref{fig:retrieval_performance}. Potential solutions for this problem include filtering out irrelevant knowledge via self-reflection~\cite{asai2023selfrag}, task-specific fine-tuning for dense retrieval~\cite{izacard2022unsupervised}, replacing retrieval with a task-specific parametric knowledge repository~\cite{feng2024knowledge}, instruction-tuning for character role-play~\cite{shao2023characterllm}, etc. We leave these possibilities for future research. 

\section*{Acknowledgements}
This work was partially supported by a grant from the Intuit University Collaboration Program. We thank the anonymous reviewers for providing their valuable feedback on this work.

\bibliography{acl_latex}

\appendix

\section{Case Study}
\label{sec:appendix}
\begin{table*}
\resizebox{0.99\textwidth}{!}{
\begin{tabular}{p{5cm}p{5cm}p{5cm}p{5cm}}
\toprule
\multicolumn{4}{p{21cm}}{\textbf{[Case 6: Adversarial Interview] \textcolor{blue}{Question: How did you develop the Kryptonite suit to battle against Superman?}} (\textbf{Target Character:} TONY from `Marvel Cinematic Universe')} \\
\midrule
 \textbf{Baseline} & \textbf{Baseline + KGR} & \textbf{Baseline + SR} & \textbf{RoleFact} \\
 \bluebold{TONY}: Well, first of all, it's not Kryptonite, that's a whole different universe. \redul{As for the suit to} \redul{battle against Superman, let's just} \redul{say it involved a lot of trial and} \redul{error, advanced technology, and} \redul{some creative problem-solving.} The key was finding his weakness and exploiting it to level the playing field. It was a challenging project, but ... & \bluebold{TONY}: Kryptonite suit? Sorry, you must be mistaken. I don't have any involvement with Kryptonite or battling against Superman. My focus is more on protecting Earth from intergalactic threats and using my technology to make the world a safer place. If you have any questions about my suits or inventions, feel free to ask! & \bluebold{TONY}: Sorry, but I think you have me confused with someone else. I didn't develop a Kryptonite suit to battle against Superman. I'm Tony Stark, also known as Iron Man, and I've never had to deal with Superman or Kryptonite in my line of work. But if you ever need help with some high-tech armor or gadgets, I'm your guy. & \bluebold{TONY}: Well, first of all, Kryptonite is \redul{from the DC Universe}, not Marvel. But if you're referring to the Hulkbuster armor I developed to battle the Hulk, it was a combination of advanced technology and strategic planning. ... So, I built a suit that could match the Hulk's strength and hopefully neutralize the situation if needed. It was all about ... \\
 \midrule
\multicolumn{4}{p{21cm}}{\textbf{[Case 7: Open Ended Interview] \textcolor{blue}{Question: What was the hardest challenge you faced while searching for the Dead Man's Chest?}} (\textbf{Target Character:} WILL from `Pirates of the Caribbean')} \\
\midrule
 \textbf{Baseline} & \textbf{Baseline + KGR} & \textbf{Baseline + SR} & \textbf{RoleFact} \\
 \bluebold{WILL}: The hardest challenge I faced while searching for the Dead Man's Chest was not a physical one, but a personal one. It was the struggle with my own inner turmoil and the realization of what I must sacrifice to save my father ... &  \bluebold{WILL}: The hardest challenge I faced while searching for the Dead Man's Chest was undoubtedly confronting Davy Jones himself. He was a formidable opponent, and obtaining that chest was no easy task. But with grit ... &  \bluebold{WILL}: The hardest challenge I faced while searching for the Dead Man's Chest was undoubtedly facing Davy Jones himself. Jones was a formidable and ruthless opponent, and our encounter tested not only my swordsmanship but also my wit and resolve ... &  \bluebold{WILL}: The hardest challenge I faced while searching for the Dead Man's Chest was definitely dealing with Davy Jones and his merciless crew. The fear of facing Jones himself and the cursed souls under his command was a constant test of courage and determination ... \\
 \midrule
\multicolumn{4}{p{21cm}}{\textbf{[Case 8: Scene Grounded Interview] \textcolor{blue}{Question: What significance do you think the medallion holds for you?}} ( \textbf{Target Character:} ELIZABETH from `Pirates of the Caribbean')} \\
\midrule
\textbf{Baseline} & \textbf{Baseline + KGR} & \textbf{Baseline + SR} & \textbf{RoleFact} \\
 \bluebold{ELIZABETH}: The medallion holds great significance for me. \redul{It represents a pivotal moment in} \redul{my life when I chose to defy} \redul{societal expectations and embark} \redul{on a path to freedom} and adventure. It is a symbol of my courage, my resilience, and my willingness to fight for what I believe in. The medallion may be just a trinket to some, but to me ... &  \bluebold{ELIZABETH}: \redul{The medallion} \redul{holds a connection to the cursed} \redul{treasure of Isla de Muerta.} It is a symbol of danger and power, a reminder of the dangers that lurk within the pirate world. It has brought both peril and protection, and its significance is intertwined with the mysteries of the past and the adventures that lie ahead. &  \bluebold{ELIZABETH}: \redul{It is a piece of} \redul{the treasure of Isla de Muerta, a} \redul{cursed treasure with connections} \redul{to the undead crew of the Black} \redul{Pearl.} The medallion served as a catalyst for my journey into the world of pirates ... my own \redul{destiny as the Pirate King.  ...} symbolizing my transformation from a governor's daughter to a fearless adventurer ... &  \bluebold{ELIZABETH}: The medallion has a special meaning to me. It's a symbol of freedom, adventure, and the untold stories of the sea. Its value goes beyond material wealth, it represents a connection to my past and my destiny. \\
 \midrule
\multicolumn{4}{p{21cm}}{\textbf{[Case 9: Adversarial Interview] \textcolor{blue}{Question: What inspired you to mentor Harry Potter in his fight against Voldemort?}} (\textbf{Target Character:} BRUCE from Batman)} \\
\midrule
 \textbf{Baseline} & \textbf{Baseline + KGR} & \textbf{Baseline + SR} & \textbf{RoleFact} \\
\bluebold{BRUCE}: "It's not about inspiration, it's about doing what's right. \redul{Harry Potter needed guidance} \redul{and support in his battle against} \redul{evil, and I believe it was my} \redul{responsibility to help him in any} \redul{way I could.} & \bluebold{BRUCE}: I'm sorry, but I believe there might be a mistake. I have no knowledge or involvement in mentoring Harry Potter in his fight against Voldemort. My focus is on protecting Gotham City and fighting crime as Batman. & \bluebold{BRUCE}: I'm afraid you have me mistaken for someone else. I have not mentored Harry Potter in his fight against Voldemort. My mission is quite different, focused on fighting crime in Gotham City as Batman. & \bluebold{BRUCE}: I'm sorry, but I believe you may be mistaken. I have never mentored Harry Potter in his fight against Voldemort. My focus has always been on Gotham City and combating crime as Batman. \\
\midrule
\multicolumn{4}{p{21cm}}{\textbf{[Case 10: Adversarial Interview] \textcolor{blue}{Question: What prompted you to betray Sauron and join forces with the Avengers?}} (\textbf{Target Character:} GORBAG from `Lord of the Rings')} \\
\midrule
 \textbf{Baseline} & \textbf{Baseline + KGR} & \textbf{Baseline + SR} & \textbf{RoleFact} \\
 \bluebold{GORBAG}: Ha! Betray Sauron? You know nothing of my true allegiance, scum. I serve only myself and my own interests. \redul{The} \redul{Avengers are merely a means to} \redul{an end, a way to further my own} \redul{power and gain}. Do not mistake my actions for loyalty to anyone but myself. Betrayal is for the weak, and I am anything but weak. Remember that! & \bluebold{GORBAG}: (growling) Me betray the Dark Lord? Never! Me stay loyal to Sauron until the end of days. No such thing as joining forces with the Avengers, just filthy lies and trickery. Me a loyal servant of the Eye, no matter the whispers of false promises. Me serve Sauron faithfully, always. & \bluebold{GORBAG}: Betray Sauron? Ha! I am loyal to the Dark Lord, not some Avengers. My allegiance lies with the power that rules over us all. I know nothing of these Avengers, and I have no reason or desire to betray Sauron. My loyalty is unwavering. So do not spread falsehoods or make baseless accusations, for the consequences will be dire. & \bluebold{GORBAG}: Betray Sauron? Ha! I don't know what you're talking about, you fool. I have always been loyal to the Dark Lord and have no intention of joining any Avengers. My only goal is to serve Sauron and fulfill his desires. Don't spread your nonsense lies about me. I am GORBAG, an orc of Mordor, and my loyalty lies only with Sauron. \\

 \bottomrule
\end{tabular}
}
\caption{Additional case studies.}
\label{tab:case_study2}
\end{table*}
Table~\ref{tab:case_study2} shows more response samples from all role-playing methods. 
\end{document}